\pdfoutput=1
\documentclass[acmsmall]{acmart}

\graphicspath{figures}

\usepackage{tabularx}
\usepackage{booktabs}
\usepackage{dcolumn}
\usepackage{hyperref}
\usepackage{amsmath}
\usepackage{amsfonts}
\usepackage{subcaption}

\usepackage{booktabs}
\usepackage{threeparttable}

\AtBeginDocument{%
  \providecommand\BibTeX{{%
    \normalfont B\kern-0.5em{\scshape i\kern-0.25em b}\kern-0.8em\TeX}}}

\setcopyright{acmcopyright}
\copyrightyear{2020}
\acmYear{2020}
\acmDOI{XX.XXXX/XXXXXXX.XXXXXXX}

\acmJournal{TOMM}
\acmVolume{0}
\acmNumber{0}
\acmArticle{0}
\acmMonth{0}

\newif\ifdraft
\draftfalse
\newcommand{\majorrevised}[1]{#1}
\newcommand{\revised}[1]{\ifdraft{\leavevmode\color{blue}{#1}}\else{#1}\fi}

\begin{document}

\title{Fine-grained Visual Textual Alignment for Cross-Modal Retrieval using Transformer Encoders}


\author{Nicola Messina}
\affiliation{%
  \institution{ISTI - CNR}
  \city{Pisa}
  \country{Italy}}
\email{nicola.messina@isti.cnr.it}

\author{Giuseppe Amato}
\affiliation{%
  \institution{ISTI - CNR}
  \city{Pisa}
  \country{Italy}}
\email{giuseppe.amato@isti.cnr.it}

\author{Andrea Esuli}
\affiliation{%
  \institution{ISTI - CNR}
  \city{Pisa}
  \country{Italy}}
\email{andrea.esuli@isti.cnr.it}

\author{Fabrizio Falchi}
\affiliation{%
  \institution{ISTI - CNR}
  \city{Pisa}
  \country{Italy}}
\email{fabrizio.falchi@isti.cnr.it}

\author{Claudio Gennaro}
\affiliation{%
  \institution{ISTI - CNR}
  \city{Pisa}
  \country{Italy}}
\email{claudio.gennaro@isti.cnr.it}

\author{Stéphane Marchand-Maillet}
\affiliation{%
  \institution{VIPER Group - University of Geneva}
  \city{Geneva}
  \country{Switzerland}}
\email{stephane.marchand-maillet@unige.ch}

\renewcommand{\shortauthors}{Messina et al.}

\begin{abstract}
 Despite the evolution of deep-learning-based visual-textual processing systems, precise multi-modal matching remains a challenging task. In this work, we tackle the task of cross-modal retrieval through image-sentence matching based on word-region alignments, using supervision only at the global image-sentence level. Specifically, we present a novel approach called Transformer Encoder Reasoning and Alignment Network (TERAN).
 TERAN enforces a fine-grained match between the underlying components of images and sentences, i.e., image regions and words, respectively, in order to preserve the informative richness of both modalities.
 TERAN obtains state-of-the-art results on the image retrieval task on both MS-COCO and Flickr30k datasets. Moreover, on MS-COCO, it also outperforms current approaches on the sentence retrieval task.
 
 Focusing on scalable cross-modal information retrieval, TERAN is designed to keep the visual and textual data pipelines well separated. 
 Cross-attention links invalidate any chance to separately extract visual and textual features needed for the online search and the offline indexing steps in large-scale retrieval systems.
 In this respect, TERAN merges the information from the two domains only during the final alignment phase, immediately before the loss computation. 
 We argue that the fine-grained alignments produced by TERAN pave the way towards the research for effective and efficient methods for large-scale cross-modal information retrieval.
 \majorrevised{We compare the effectiveness of our approach against relevant state-of-the-art methods. On the MS-COCO 1K test set, we obtain an improvement of 5.7\% and 3.5\%} respectively on the image and the sentence retrieval tasks on the Recall@1 metric.
 The code used for the experiments is publicly available on GitHub at \url{https://github.com/mesnico/TERAN}.
\end{abstract}

\ccsdesc[300]{Computing methodologies~Semi-supervised learning settings}
\ccsdesc[500]{Information systems~Multimedia and multimodal retrieval}
\ccsdesc[500]{Computing methodologies~Neural networks}

\keywords{deep learning, cross-modal retrieval, multi-modal matching, computer vision, natural language processing}

\maketitle

\section{Introduction}

Since 2012, deep learning has obtained impressive results in several vision and language tasks. Recently, various attempts have been made to merge the two worlds, and state-of-the-art results have been obtained in many of these tasks, including visual question answering \cite{hu2017learning,Anderson2018bottomup,teney2017graph}, image captioning \cite{zhou2020unified,rennie2017self,huang2019attention,cornia2019show}, and image-text matching \cite{chen2019uniter,lu2019vilbert,vsepp2018faghri,lee2018stackedcrossattention}. In this work, we deal with the cross-modal retrieval task, with a focus on the visual and textual modalities. 
The task consists in finding the top-relevant images representing a natural language sentence given as a query (image-retrieval), or, vice versa, in finding a set of sentences that best describe an image given as a query (sentence-retrieval).

This cross-modal retrieval task is closely related to image-sentence matching, which consists in assigning a score to a pair composed of an image and a sentence. The score is high if the sentence adequately describes the image, and low if the input sentence is unrelated to the corresponding image. The score function learned by solving the matching problem can be then used for deciding which are the top-relevant images and sentences in the two image- and sentence- retrieval scenarios. 
The matching problem is often very difficult since a deep high-level understanding of images and sentences is needed for succeeding in this task. 

Visuals and texts are used by humans to fully understand the real world. Although they are of equal importance, the information hidden in these two modalities has a very different nature. The text is already a well-structured description developed by humans in hundreds of years, while images are nothing but raw matrices of pixels hiding very high-level concepts and structures. 
Images and texts do not describe only static entities. In fact, they can easily portray relationships between the objects of interest, e.g.: "The kid \textit{kicks} the ball". Therefore, it would be helpful to also understand spatial and even abstract relationships linking them together. 


Vision and language matching has been extensively studied \cite{vsepp2018faghri,carrara2018pictureit,lu2019vilbert,karpathy2015alignment,lee2018stackedcrossattention}.
Many works employ standard architectures for processing images and texts, such as CNNs-based models for image processing and recurrent networks for language.
Usually, in this scenario, the image embeddings are extracted from standard image classification networks, such as ResNet or VGG, by employing the network activations before the classification head. Usually, descriptions extracted from CNN networks trained on classification tasks can only capture global summarized features of the image, ignoring important localized details. \majorrevised{For this reason, recent works make extensive use of attention mechanisms, which are able to relate each visual object, extracted from the spatial locations of a feature map or an object detector to the most interesting parts of the sentence, and/or vice-versa.

\revised{Many of these works, such as ViLBERT\cite{lu2019vilbert}, ImageBERT\cite{qi2020imagebert}, VL-BERT\cite{Su2020VL-BERT}, IMRAM\cite{Chen2020imram}, try to learn a complex scoring function $s = \phi(I, C)$ that measures the affinity between an image and a caption, where $I$ is an image, $C$ is the caption and $s$ is a normalized score in the range $[0, 1]$. These are very effective models for tackling the matching task, and they reach state-of-the-art results. However, they remain very inefficient for large-scale image or sentence retrieval: the problem with these approaches is that it is not possible to extract visual and textual descriptions separately, as the pipelines are strongly entangled through cross-attention or memory layers. Thus, if we want to retrieve images related to a given query text, we have to compute all the similarities using the $\phi$ function and then sort the resulting scores in descending order. This is unfeasible if we want to retrieve images or sentences from a large database in a few milliseconds.}}

In our previous work, we introduced the Transformer Encoder Reasoning Network (TERN) architecture \cite{messina2020tern}, which is a transformer-based model able to independently process images and sentences to match them into the same common space. TERN is a useful architecture for producing compact yet informative features that could be used in cross-modal retrieval setups for efficient indexing using metric-space or text-based approaches.
\majorrevised{TERN processes visual and textual elements using transformer encoder layers, exploring and reasoning on the relationships among image regions and sentence words. However, its main objective is to match images and sentences as atomic, global entities, by learning a global representation of them inside special tokens (I-CLS and T-CLS) processed by the transformer encoder.} 
This usually leads to performance loss \majorrevised{and possibly poor generalization} since fine-grained information useful for effective matching is lost during the projection to a fixed-sized common space.

\majorrevised{For this reason, in this work, we propose TERAN (Transformer Encoder Reasoning and Alignment Network) in which we force a fine-grained word-region alignment. Fine-grained matching deals with the accurate understanding of the local correspondences between image regions and words, as opposed to coarse-grained matching, where only a summarized global descriptions of the two modalities is considered.
In fact, differently from TERN, the objective function is directly defined on the set of regions and words in output from the architecture, and not on a potentially lossy global representation. Using this objective, TERAN tries to individually align the regions and the words contained in images and sentences respectively, instead of directly matching images and sentences as a whole. The information available to TERAN during training is still coarse-grained, as we do not inject any information about word-region correspondences. The fine-grained alignment is thus obtained in a semi-supervised setup, where no explicit word-region correspondences are given to the network.}

Our TERAN proposal shares most of the previous TERN building blocks and interconnections: the visual and textual pipelines are forwarded separately and they are fused only during the loss computation, in the very last stage of the architecture, making scalable cross-modal information retrieval possible. At the same time, this novel architecture employs state-of-the-art self-attentive modules, based on the transformer encoder architecture \cite{vaswani2017transformer}, able to spot out hidden relationships in both modalities for a very effective fine-grained alignment. 


\majorrevised{Therefore, TERAN is able to produce independent visual and textual features usable in efficient retrieval scenarios implementing two simple visual and textual pipelines built of modern self-attentive mechanisms. In spite of its overall simplicity, TERAN is able to reach state-of-the-art results in the image and sentence retrieval task, even when compared with complex entangled visual-textual matching models. Experiments show that TERAN can generalize better with respect to the previous TERN approach.}

In the evaluation of the proposed matching procedure, we used a typical information retrieval setup using the Recall@K metrics (with $K = \{1, 5, 10\}$.)
However, in common search engines where the user is searching for related images and not necessarily exact matches, the Recall@K evaluation could be too rigid, especially when $K = 1$.
For this reason, as in our previous work \cite{messina2020tern}, in addition to the strict Recall@K metric, we propose to measure the retrieval abilities of the system with a normalized discounted cumulative gain metric (NDCG) with relevance computed exploiting caption similarities.



Summarizing, the contributions of this paper are the following:
\begin{itemize}
    \item we introduce the Transformer Encoder Reasoning and Alignment Network (TERAN), able to produce fine-grained region-word alignments for efficient cross-modal information retrieval.
    \item we show that TERAN can reach state-of-the-art results on the cross-modal visual-textual retrieval task, both in terms of Recall@K and NDCG, while producing visually-pleasant region-words alignments without using supervision at the region-word level. Retrieval results are measured both on MS-COCO and Flickr30k datasets.
    \item \majorrevised{we quantitatively compare TERAN with our previous work \cite{messina2020tern}, and we perform an extensive study on several variants of our novel model, including weight sharing in the last transformer layers, stop-words removal during training, different pooling protocols for the matching loss function, and the usage of different language models}.
\end{itemize}

\section{Related Work}
In this section, we review some of the previous works related to image-text joint processing for cross-modal retrieval \majorrevised{and alignment}, and high-level relational reasoning, on which this work lays its foundations. Also, we briefly summarize the evaluation metrics available in the literature for the cross-modal retrieval task.

\subsection*{Image-Text Processing for Cross-Modal Retrieval}
Image-text matching is often cast to the problem of inferring a similarity score among an image and a sentence. Usually, one of the common approaches for computing this cross-domain similarity is to project images and texts into a common representation space on which some kind of similarity measure can be defined (e.g.: cosine or dot-product similarities).
Images and sentences are preprocessed by specialized architectures before being merged at some point in the pipeline.

Concerning image processing, the standard approach consists in using Convolutional Neural Networks (CNNs), usually pre-trained on image classification tasks. In particular, \cite{KleinLSW15fishervectors,VendrovKFU15,LinP16,HuangWW17,EisenschtatW17} use VGGs, while \cite{LiuGBL17,vsepp2018faghri,GuCJN018,Huang2018} use ResNets. 
\majorrevised{
Concerning sentence processing, many works \cite{karpathy2015alignment,vsepp2018faghri,li2019,lee2018stackedcrossattention,Huang2018} employ GRU or LSTM recurrent networks to process natural language, often considering the final hidden state as the only feature representing the whole sentence.
The problem with these kinds of methodologies is that they usually extract extremely summarized global descriptions of images and sentences.
Therefore, a lot of useful fine-grained information needed to reconstruct inter-object relationships for precise image-text alignment is permanently lost.

For these reasons, many works try to employ region-level information, together with word-level descriptions provided by recurrent networks, to understand fine-grained alignments between words and localized patches in the image.
Recent works \cite{huang2018bi,liu2019focus,liu2020graph,huang2019acmm,wang2019position,Chen2020imram,lee2018stackedcrossattention} exploit the availability of pre-computed region-level features extracted from the Faster-RCNN \cite{RenHGS15fasterrcnn} object detector. An alternative consists in using the features maps in output from ResNets, without aggregating them, for computing fine-grained attentions over the sentences \cite{xu2020cross,huang2018image,wang2018joint,wei2020adversarial,guo2020associating,ji2020sman}.


Recently, the transformer architecture \cite{vaswani2017transformer} achieved state-of-the-art results in many natural language processing tasks, such as next sentence prediction or sentence classification. The results achieved by the BERT model \cite{devlin2019bert} are a demonstration of the power of the attention mechanism to produce accurate context-aware word descriptions.
For this reason, some works in image-text matching use BERT to extract contextualized word embeddings for representing sentences \cite{wu2019learning,sarafianos2019adversarial,qu2020context,wei2020multi}. \revised{Drawing inspiration from the powerful contextualization capabilities of the transformer encoder architecture, some works use BERT-like processing on both visual and textual modalities, such as ViLBERT \cite{lu2019vilbert}, ImageBERT \cite{qi2020imagebert}, Pixel-BERT \cite{huang2020pixelbert}, VL-BERT \cite{Su2020VL-BERT}. 

These latest works achieve state-of-the-art results in sentence and image retrieval, as well as excellent results on the downstream word-region alignment task \cite{chen2019uniter}. However, they cannot produce separate image and caption descriptions; this is an important requirement in real-world search engines, where usually, at query time, only the query element is forwarded through the network, while all the elements of the database have already been processed by means of an offline feature extraction process.}


Some architectures have been designed so that they are natively able to extract disentangled visual and textual features. In particular, } in \cite{vsepp2018faghri} the authors introduce the VSE++ architecture. They use VGG and ResNets visual features extractors, together with an LSTM for sentence processing, and they match images and captions exploiting hard-negatives during the loss computation. 
With their VSRN architecture \cite{li2019}, the authors introduce a visual reasoning pipeline built of Graph Convolution Networks (GCNs) and a GRU to sequentially reason on the different image regions. Furthermore, they impose a sentence reconstruction loss to regularize the training process. 
\majorrevised{The authors in \cite{huang2018image} use a similar objective, but employing a pre-trained multi-label CNN to find semantically relevant image patches and their vectorial descriptions.
Differently, in \cite{sarafianos2019adversarial} an adversarial learning method is proposed, where a discriminator is used to learn modality-invariant representations. The authors in \cite{guo2020associating} use a contextual attention-based LSTM-RNN which can selectively attend to salient regions of an image at each time step, and they employ a recurrent canonical correlation analysis to find hidden semantic relationships between regions and words.

The works closer to our setup are SAEM \cite{wu2019learning} and CAMERA \cite{qu2020context}. In \cite{wu2019learning} the authors use triplet and angular loss to project the image and sentence features into the same common space. The visual and textual features are obtained through transformer encoder modules. Differently from our work, they do not enforce fine-grained alignments and they pool the final representations to obtain a single-vector representation. Instead, in \cite{qu2020context} the authors use BERT as language model and an adaptive gating self-attention module to obtain context-enhanced visual features, projecting them into the same common space using cosine similarity. Unlike our work, they specifically focus on multi-view summarization, as multiple sentences can describe the same images in many different but complementary ways.



The loss used in our work is inspired by the matching loss introduced by the MRNN architecture \cite{karpathy2015alignment}, which seems able to produce very good region-word alignments by supervising only the global image-sentence level.
}

\subsection*{High-Level Reasoning}
Another branch of research from which this work draws inspiration is focused on the study of relational reasoning models for high-level understanding. The work in \cite{santoro2017rn} proposes an architecture that separates perception from reasoning. They tackle the problem of Visual Question Answering by introducing a particular layer called Relation Network (RN), which is specialized in comparing pairs of objects. Object representations are learned using a four-layer CNN, and the question embedding is generated through an LSTM. The authors in \cite{messina2019avfrn,DBLP:messina2019cbir} extend the RN for producing compact features for relation-aware image retrieval. However, they do not explore the multi-modal retrieval setup.

Other solutions try to stick more to a symbolic-like way of reasoning. In particular, \cite{hu2017learning,inferring_and_executing_programs} introduce compositional approaches able to explicitly model the reasoning process by dynamically building a reasoning graph that states which operations must be carried out and in which order to obtain the right answer.

Recent works employ Graph Convolution Networks (GCNs) to reason about the interconnections between concepts. The authors in \cite{YaoPLM18,YangTZC19,LiJ19} use GCNs to reason on the image regions for image captioning, while \cite{YangLLBP18graphrcnn,LiOZSZW18} use GCN with attention mechanisms to produce the scene graph from plain images.

\subsection*{Cross-Modal Retrieval Evaluation Metrics}
All the works involved with image-caption matching evaluate their results by measuring how good the system is at retrieving relevant images given a query caption (image-retrieval) and vice-versa (caption-retrieval). 

Usually the Recall@K metric is used \cite{vsepp2018faghri,li2019,qi2020imagebert,lu2019vilbert,lee2019}, where typically $K = \{1, 5, 10\}$.
On the other hand, \cite{carrara2018pictureit} introduced a novel metric able to capture non-exact results by weighting the ranked documents using a caption-based similarity measure.

We extend the metric introduced in \cite{carrara2018pictureit}, giving rise to a powerful evaluation protocol that handles non-exact yet relevant matches. Relaxing the constraints of exact-match similarity search is an important step towards an effective evaluation of real search engines.

\section{Review of Transformer Encoders}
\label{sec:te}
Our proposed architecture is based on the well established Transformer Encoder (TE) architecture, which heavily relies on the concept of self-attention.
The self-attention mechanism tries to weight every vector of the sequence using a scalar value normalized in the range $[0, 1]$ computed as a function of the input vectors themselves. In particular, the attention is computed by using a \textit{query} vector $Q$ and a set of \textit{key-value} ($K$, $V$) pairs derived from data using simple feed-forward networks, and processed as shown in Equation \ref{eq:te}. More in detail, the attention-aware vector in output from the attention module is computed for every input element as a weighted sum of the values, where the weight assigned to each value is computed as a similarity score (scaled dot-product) between the query with the corresponding key:



\begin{equation}
\label{eq:te}
    \text{Att}(Q, K, V) = \text{softmax} \left( \frac{QK^T}{\sqrt{d_k}}\right)V.
\end{equation}
$Q, K, V$ are the query, the key, and the value respectively, while the factor $\sqrt{d_k}$ is used to mitigate the vanishing gradient problem of the softmax function in case the inner product assumes too large values. \majorrevised {In real implementations, a multi-head attention is used: the input vectors are chunked, and every chunk is processed independently using a different instantiation of the above-described mechanism. This helps in capturing the relationships between the different portions of every input vector.}



\begin{figure*}[t]
    \centering
    \includegraphics[page=7, width=0.6\linewidth]{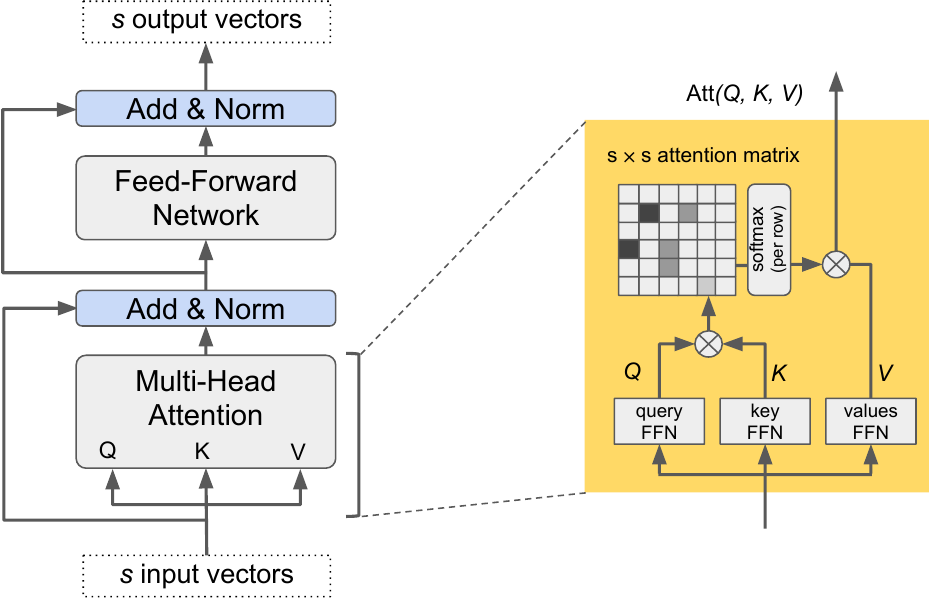}
  \caption{\majorrevised{A high-level view of the transformer encoder layer. Every arrow carries $s$ fixed-sized vectors.}}
  \label{fig:transformer_encoder} 
\end{figure*}

Finally, the output from the TE is computed through a simple feed-forward layer applied to the $Att(Q, K, V)$ vectors, with a ReLU activation function. This simple feed-forward layer casts in output a set of features having the same dimensionality of the input sequence. Two residual connections followed by layer normalization are also present around the self-attention and the feed-forward sub-modules. \majorrevised{An overview of the transformer encoder architecture is shown in Figure~\ref{fig:transformer_encoder}}.

Although the TE was initially developed to work on sequences, there are no architectural constraints that prevent its usage on sets of vectors instead of sequences. In fact, the TE module has not any built-in sequential prior which considers every vector in a precise position in the sequence. This makes the TE suitable for processing visual features coming from an object detector.

We argue that the transformer encoder self-attention mechanism can drive a simple but powerful reasoning mechanism able to spot hidden relationships between the vector entities, whatever nature they have (visual or textual). Also, the encoder is designed in a way that multiple instances of it could be stacked in sequence. Using multiple levels of attention helps in producing a deeper and more powerful reasoning pipeline.



\section{Transformer Encoder Reasoning and Alignment Network (TERAN)}
Our Transformer Encoder Reasoning and Alignment Network (TERAN) leverages our previous work \cite{messina2020tern} that introduced the TERN architecture. TERAN modifies the learning objective of our previous work by forcing a fine-grained alignment between the region and word features in output from the last transformer encoder (TE) layers so that meaningful fine-grained concepts are produced. 

As TERN, our TERAN reasoning engine is built using a stack of TE layers, both for the visual and the textual data pipelines. The TE takes as input sequences or sets of entities, and it can reason upon these entities disregarding their intrinsic nature.
In particular, we consider the salient regions in an image as visual entities, and the words present in the caption as textual entities.

More formally, the input to our reasoning pipeline is a set $I = \{r_0, r_1, \ldots, r_n\}$ of $n$ image regions (visual entities) representing an image $I$ and a sequence $C = \{w_0, w_1, \ldots, w_m\}$ of $m$ words (textual entities) representing the corresponding caption $C$.
Thus, the reasoning module continuously operates on sets and sequences of $n$ and $m$ objects respectively for images and captions. 

The TERN architecture in \cite{messina2020tern} produces summarized representations of both images and words by employing special I-CLS and T-CLS tokens that are forwarded towards the layers of the TEs. In the end, the processed I-CLS and T-CLS tokens gather important global knowledge from both modalities.
Contrarily, TERAN does not produce aggregated fixed-sized representations for images and sentences. \revised{For this reason, it does not employ the global features constructed inside the I-CLS and T-CLS tokens}. Instead, it tries to impose a global matching loss defined on the variable-length sets in output from the last TE layers that is able, as a side effect, to produce also good and interpretable region-word alignments.





\begin{figure*}[t]
    \centering
    \includegraphics[page=1, width=1\linewidth]{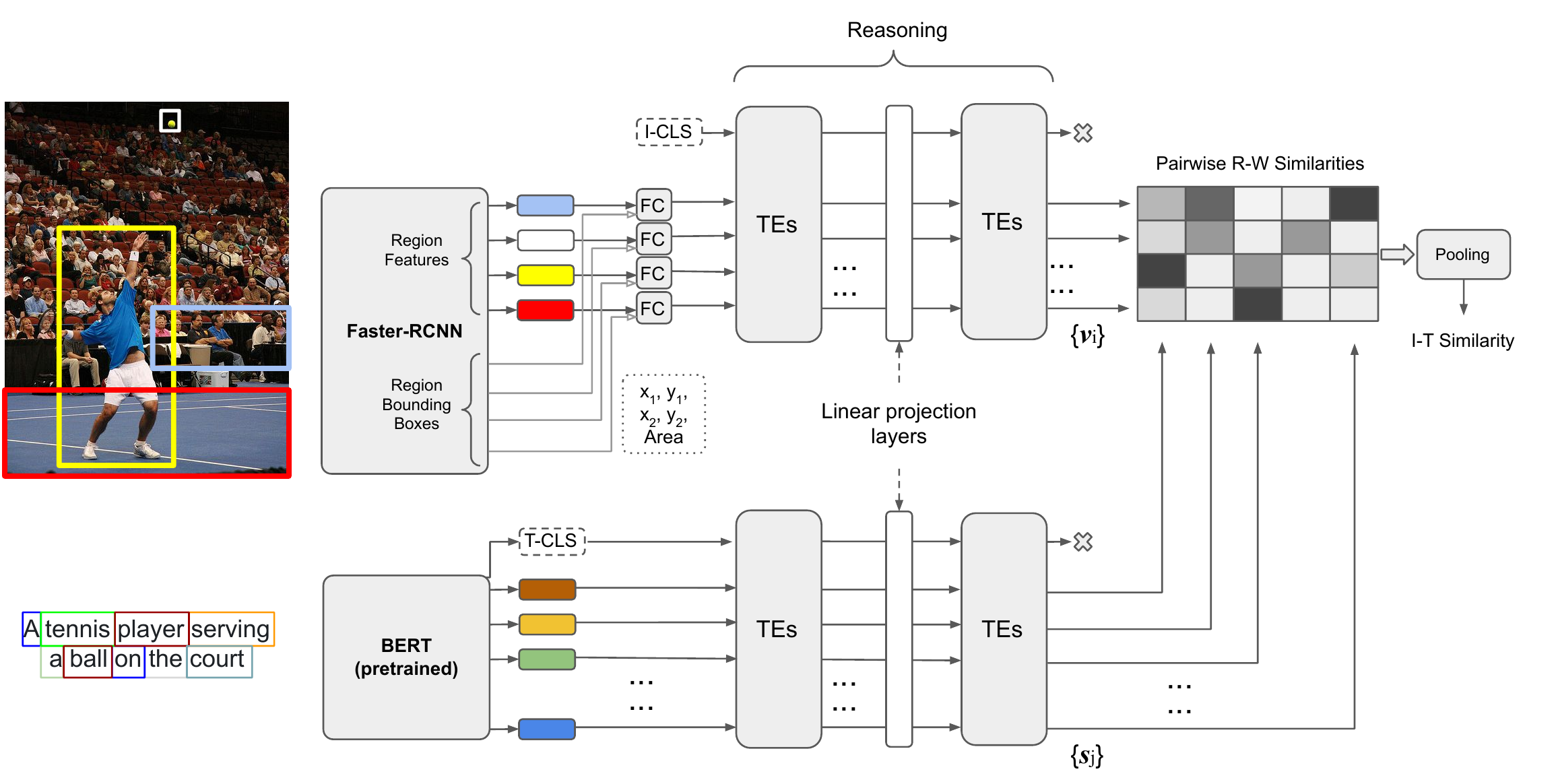}
  \caption{The proposed TERAN architecture. \revised{TEs stands for Transformer Encoders, and it indicates a stack of TE layers whose internals are recalled in Section \ref{sec:te} and explained in detail in \cite{vaswani2017transformer}. Region and word features are extracted through a bottom-up attention model based on Faster-RCNN and BERT, respectively. The final image-text (I-T) similarity score is obtained by pooling a region-word (R-W) similarity matrix. Note that the special I-CLS and T-CLS are not used in the basic formulation of TERAN.}}
  \label{fig:detailed_architecture} 
\end{figure*}

The overall architecture is shown in Figure \ref{fig:detailed_architecture}. \revised{We left in the scheme the I-CLS and T-CLS tokens connections for comparison with the TERN architecture presented in \cite{messina2020tern}. These tokens are still used for a targeted experiment that exploits the combination of the TERN and TERAN losses (more details in Section \ref{sec:experiments}). However, they are not used in the main TERAN experiments.}

\revised{The visual features extracted from Faster-RCNN are conditioned with the information related to the geometry of the bounding-boxes. This is done through a simple fully-connected stack in the early visual pipeline before the reasoning steps.}
The two linear projection layers within the TE modules are used to project the visual and textual concepts in spaces having the same dimensionality. Then, the latest TE layers perform further processing before outputting the final features that are used to compute the final matching loss.

Differently from TERN, we initially do not share the weights of the last TE layers. We will discuss the effect of weight sharing in our ablation study, in Section \ref{sec:weight_sharing}.

In our novel TERAN architecture, the features in output from the last TE layers are used to compute a region-word alignment matrix $A \in \mathbb{R}^{|g_k| \times |g_l|}$, where $g_k$ is the set of indexes of the region features from the $k$-th image and $g_l$ is the set of indexes of the words from the $l$-th sentence. We use cosine similarity for measuring affinity between the $i$-th region and the $j$-th word. 
If $\{\boldsymbol{v}_i\}$ and $\{\boldsymbol{s}_j\}$ are the sets of contextualized region and word vectors in output from the network for the k-th image and the l-th sentence respectively, then $A$ is constructed as:
\begin{equation}
    A_{ij} = \frac{\boldsymbol{v}_i^T \boldsymbol{s}_j}{\| \boldsymbol{v}_i \| \| \boldsymbol{s}_j \|} \qquad i \in g_k, j \in g_l
\end{equation}

At this point, the global similarity $S_{kl}$ between the $k$-th image and the $l$-th sentence is computed by pooling this similarity matrix through an appropriate pooling function. Inspired by \cite{karpathy2015alignment} and \cite{lee2018stackedcrossattention}, we employ the max-sum pooling, which consists in computing the max over the rows of $A$ and then summing or, equivalently, \textit{max-over-regions sum-over-words} ($M_{r}S_{w}$) pooling. We explore also the dual version, as in \cite{lee2018stackedcrossattention}, by computing the max over the columns and then summing, or \textit{max-over-words sum-over-regions} ($M_{w}S_{r}$) pooling:

\begin{equation}
    S^{M_{r}S_{w}}_{kl} = \sum_{j \in g_l}\max_{i \in g_k} {A_{ij}} \qquad or \qquad S^{M_{w}S_{r}}_{kl} = \sum_{i \in g_k}\max_{j \in g_l} {A_{ij}}
\end{equation}

Since both these similarity functions are not symmetric due to the diverse outcomes we obtain by inverting the order of the sum and max operations, we introduce also the symmetric form, obtained by summing the two:
\begin{equation}
    S^{\text{Symm}}_{kl} = S^{M_{r}S_{w}}_{kl} + S^{M_{w}S_{r}}_{kl}
\end{equation}

\subsection{Learning Objective}
Given the global image-sentence similarities $S_{kl}$ computed through alignments pooling, we can proceed as in previous works \cite{vsepp2018faghri,li2019} using a contrastive learning method: we use a hinge-based triplet ranking loss, focusing the attention on hard negatives, as introduced by \cite{vsepp2018faghri}.
Therefore, we used the following loss function:
\begin{equation}
\begin{split}
    L_{kl} &= \max_{{l}'} [\alpha + S_{kl'} - S_{kl}]_+ + \\
    &\mathrel{\phantom{=}} \max_{{k}'} [\alpha + S_{k'l} - S_{kl}]_+
\end{split}
\end{equation}

where $[x]_+ \equiv max(0, x)$ and $\alpha$ is a margin that defines the minimum separation that should hold between the truly matching word-region embeddings and the negative pairs. The hard negatives ${k}'$ and ${l}'$ are computed as follows: 
\begin{equation}
\begin{split}
    {k}' = \text{arg} \max_{j \neq k} S(j, l) \\
    {l}' = \text{arg} \max_{d \neq l} S(k, d)
\end{split}
\end{equation}

where $(k, l)$ is a positive pair.
As in \cite{vsepp2018faghri}, the hard negatives are sampled from the mini-batch and not globally, for performance reasons.

\subsection{Region and Word Features Extraction}
The $I = \{r_0, r_1, \ldots, r_n\}$ and $C = \{w_0, w_1, \ldots, w_m\}$ initial descriptions for images and captions come from state-of-the-art visual and textual pre-trained networks, Faster-RCNN with Bottom-Up attention and BERT respectively.

Faster-RCNN \cite{RenHGS15fasterrcnn} is a state-of-the-art object detector. It has been used in many downstream tasks requiring salient object regions extracted from images. 
Therefore, Faster-RCNN is one of the main architectures implementing human-like visual perception.
The work in \cite{Anderson2018bottomup} introduces bottom-up visual features by training Faster-RCNN with a Resnet-101 backbone on the Visual Genome dataset \cite{Krishna2016VisualGenome}. Using these features, they can reach remarkable results on the two downstream tasks of image captioning and visual question answering.


Concerning text processing, we use BERT \cite{devlin2019bert} for extracting word embeddings. BERT already uses a multi-layer transformer encoder to process words in sentences and capture their functional relationships through the same powerful self-attention mechanism. BERT embeddings are trained on some general natural language processing tasks such as sentence prediction or sentence classification and demonstrated state-of-the-art results in many downstream natural language tasks.
BERT embeddings, unlike word2vec \cite{Mikolov2013word2vec}, capture the context in which each word appears. Therefore, every word embedding carries information about the surrounding context, that could be different from caption to caption.

Since the transformer encoder architecture does not embed any sequential prior in its architecture, words are given a sequential order by mixing some positional information into the learned input embeddings. For this reason, the authors in \cite{vaswani2017transformer} add sine and cosine functions of different frequencies to the input embeddings. This is a simple yet effective way to transform a set into a sequence.

\majorrevised{
\section{Computational Efficiency of TERAN}
\label{sec:computational_efficiency}
A principled objective of our work is efficient feature extraction for cross-modal retrieval applications. In these scenarios, it is mandatory to have a separable network that can produce visual and textual features by independently forwarding the two disentangled visual and textual pipelines. Furthermore, the similarity function should be simple, so that it is efficient to compute.

TERAN, as well as TERN \cite{messina2020tern} and other works in literature \cite{wu2019learning,qu2020context,li2019} adhere to this principle. In fact, if $K$ is the number of images and $L$ the number of sentences in the database, these methods have a feature space complexity, as well as a feature extraction time complexity, of $O(K) + O(L)$.

Other works that entangle the visual and textual pipelines, such as \cite{Chen2020imram,xu2020cross,wang2019position} require a feature space, and a number of network evaluations, scaling with $O(KL)$. These methods are impractical to deploy to real-world scalable search engines. Some of these works partially solve this issue by keeping the two representations separated up until a certain point in the network, so that these intermediate representations can be cached, as proposed in \cite{macavaney2020efficient}.
In all these cases, a new incoming query to the system needs $O(K)$ or $O(L)$ re-evaluations of the whole network (depending on whether we are considering image or sentence retrieval); in the best case, we need to re-evaluate the last attention layers, which could be similarly expensive.

Regarding the similarity computation, TERN uses simple dot products that enable quick and efficient document rankings in modern search engines.
TERAN implements also a very simple similarity function, built of simple dot products and summations without including complex layers of memories or attentions. This possibly enables an implementation that uses metric space approaches to prune the search space and obtain very efficient image or sentence rankings for a given query. However, the implementation of the TERAN similarity function in real-world search engines is left for future research.
}

\section{NDCG metric for Cross-Modal Retrieval}
\label{sec:computing_rel}
As of now, many works in the computer vision literature treating image-text matching measure the retrieval abilities of the proposed methods by employing the well known Recall@K metric. The Recall@K measures the percentage of queries able to retrieve the correct item among the first K results. 
This is a metric perfectly suitable for scenarios where the query is very specific and thus we expect to find the elements that match perfectly among the first search results.
However, in common search engines, the users are not asked to input a very detailed query, and they are often not searching for an exact match. They expect to find in the first retrieved positions some relevant results, with relevance defined using some pre-defined and often subjective criterion.


For this reason, inspired by the work in \cite{carrara2018pictureit} and following the novel ideas introduced by our previous work on TERN \cite{messina2020tern}, we employ a common metric often used in information retrieval applications, the Normalized Discounted Cumulative Gain (NDCG).
The NDCG is able to evaluate the quality of the ranking produced by a certain query by looking at the first $p$ positions of the ranked elements list. 
The premise of NDCG is that highly relevant items appearing lower in a search result list should be penalized as the graded relevance value is reduced proportionally to the position of the result.

The NDCG until position $p$ is defined as follows:
\begin{equation}
    \text{NDCG}_{p} = \frac{\text{DCG}_p}{\text{IDCG}_p}, \qquad \text{where} \quad DCG_{p} = \sum _{i=1}^{p}{\frac {\text{rel}_{i}}{\log _{2}(i+1)}};
\end{equation}

$\text{rel}_i$ is a positive number encoding the affinity that the $i$-th element of the retrieved list has with the query element, and $\text{IDCG}_p$ is the $\text{DCG}_p$ of the best possible ranking. Thanks to this normalization, $\text{NDCG}_p$ acquires values in the range $[0, 1]$. 


The $\text{rel}_i$ values can be computed using well-established sentence similarity scores between a sentence and the sentences associated with a certain image.
More formally, we could think of computing $\text{rel}_i = \tau(\bar{C}_i, C_j)$, where $\bar{C}_i$ is the set of all captions associated to the image $I_i$, and $\tau : \mathbb{S} \times \mathbb{S} \rightarrow [0, 1] $ is a similarity function defined over a pair of sentences returning their normalized similarity score.
With this simple expedient, we could efficiently compute quite large relevance matrices using similarities defined over captions, which are in general computationally much cheaper than similarities computed between images and sentences directly.


We thus compute the $rel_i$ value in the following ways:
\begin{itemize}
    \item $\text{rel}_i = \tau(\bar{C}_i, C_j)$ in case of image retrieval, where $C_j$ is the query caption
    \item $\text{rel}_i = \tau(\bar{C}_j, C_i)$ in case of caption retrieval, where $\bar{C}_j$ is the set of captions associated to the query image $I_j$. 
\end{itemize}

In our work, we use \texttt{ROUGE-L}\cite{lin-2004-rouge} and \texttt{SPICE}\cite{AndersonFJG16spice} as sentence similarity functions $\tau$ for computing caption similarities. These two scoring functions capture different aspects of the sentences. In particular, \texttt{ROUGE-L} operates on the longest common sub-sequences, while \texttt{SPICE} exploits graphs associated with the syntactic parse trees, and has a certain degree of robustness against synonyms. In this way, \texttt{SPICE} is more sensitive to high-level features of the text and semantic dependencies between words and concepts rather than to pure syntactic constructions.

\section{Experiments}
\label{sec:experiments}
We trained the TERAN architecture and we measured its performance on the MS-COCO \cite{LinMBHPRDZ14coco} and Flickr30k datasets \cite{young2014image}, computing the effectiveness of our approach on the image retrieval and sentence retrieval tasks. We compared our results against state-of-the-art approaches on the same datasets, using the introduced NDCG and the already-in-use Recall@K metrics.

The MS-COCO dataset comes with a total of 123,287 images. Every image has associated a set of 5 human-written captions describing the image. 
We follow the splits introduced by \cite{karpathy2015alignment} and followed by the subsequent works in this field \cite{vsepp2018faghri,GuCJN018,li2019}. In particular, 113,287 images are reserved for training, 5,000 for validating, and 5,000 for testing.
Differently, Flickr30k consists of 31,000 images and 158,915 English texts. Like MS-COCO, each image is annotated with 5 captions. Following the splits by \cite{karpathy2015alignment}, we use 29,000 images for training, 1,000 images for validation, and the remaining 1,000 images for testing.
For MS-COCO, at test time the results for both 5k and 1k test-sets are reported. In the case of 1k images, the results are computed by performing 5-fold cross-validation on the 5k test split and averaging the outcomes.

We computed caption-caption relevance scores for the NDCG metric using \texttt{ROUGE-L}\cite{lin-2004-rouge} and \texttt{SPICE}\cite{AndersonFJG16spice}, as explained in Section \ref{sec:computing_rel}, and we set the NDCG parameter $p = 25$ as in \cite{carrara2018pictureit} in our experiments.
We employed the NDCG metrics measured during the validation phase for choosing the best performing model to be used during the test phase.

\majorrevised{For a better comparison with our previous TERN approach, we included three more targeted experiments.}
In the first two, called \textit{TERN $M_{w}S_{r}$ Test} and \textit{TERN $M_{r}S_{w}$ Test} we used the best-performing TERN model, trained as explained in \cite{messina2020tern}, testing it using the $M_{w}S_{r}$ and $M_{r}S_{w}$ alignments criteria respectively. TERN is effectively able to output features for every image region or word; however, it is never constrained to produce meaningful descriptions out of these sets of features; hence, this trial is aimed at checking the quality of the alignment of the concepts in output from the previous TERN architecture. 
\majorrevised{In the third experiment, called \textit{TERN w. Align}, we tried to integrate the objectives of both TERN and TERAN during training, by combining their losses using the uncertainty weighting method proposed in \cite{kendall2018multi}, and testing the model using the TERN inference protocol. \revised{Thus, in this experiment, we effectively reuse the I-CLS and T-CLS tokens as global descriptions for images and sentences, as described in \cite{messina2020tern}}. This experiment aimed to evaluate if the TERAN alignment objective can help TERN learn better fixed-sized global vectorial descriptions.
}

\subsection{Implementation Details}
We employ the BERT model pre-trained on the masked language task on English sentences, using the PyTorch implementation by HuggingFace \footnote{\url{https://github.com/huggingface/transformers}}. These pre-trained BERT embeddings are 768-D.
For the visual pipeline, we extracted the bottom-up features from the work by \cite{Anderson2018bottomup}, using the code and pre-extracted features provided by the authors \footnote{\url{https://github.com/peteanderson80/bottom-up-attention}}. Specifically, for MS-COCO we used the already-extracted bottom-up features, while we extracted from scratch the features for Flickr30k using the available pre-trained model.

In the experiments, we used the bottom-up features containing the top 36 most confident detections, although our pipeline already handles variable-length sets of regions for each image by appropriately masking the attention weights in the TE layers.

Concerning the reasoning steps, we used a stack of 4 TE layers for visual reasoning. We found the best results when fine-tuning the BERT pre-trained model, so we did not add further reasoning TE layers for the textual pipeline.
The final common space, as in \cite{vsepp2018faghri}, is 1024-dimensional. We linearly projected the visual and textual features to a 1024-d space and then we processed the resulting features using 2 final TEs before computing the alignment matrix.

All the TEs feed-forward layers are 2048-dimensional and the dropout is set to 0.1.
We trained for 30 epochs using Adam optimizer with a batch size of 40 and a learning rate of $1e{-5}$ for the first 20 epochs and $1e{-6}$ for the remaining 10 epochs.
The $\alpha$ parameter of the hinge-based triplet ranking loss is set to 0.2, as in \cite{vsepp2018faghri,li2019}.


\subsection{Results}
\majorrevised{We compare our TERAN method against the following baselines: JGCAR \cite{wang2018joint}, SAN \cite{ji2019saliency}, VSE++ \cite{vsepp2018faghri}, SMAN \cite{ji2020sman}, M3A-Net \cite{ji2020multi}, AAMEL \cite{wei2020adversarial}, MRNN \cite{karpathy2015alignment}, SCAN \cite{lee2018stackedcrossattention}, SAEM \cite{wu2019learning}, CASC \cite{xu2020cross}, MMCA \cite{wei2020multi}, VSRN \cite{li2019}, PFAN \cite{wang2019position}, Full-IMRAM \cite{Chen2020imram}, and CAMERA \cite{qu2020context}.
We clustered these methods based on the visual feature extractor they use: VGG, ResNet, or Region CNN (e.g., Faster-RCNN). To have a better comparison with our method, we also annotated in the tables whenever they use BERT as the textual model, or if they use disentangled visual-textual pipelines for efficient feature computation, as explained in Section \ref{sec:computational_efficiency}.
Also note that many of the listed methods report the results using an ensemble of two models having different training initialization parameters, where the final similarity is obtained by averaging the scores in output from each model. Hence, we reported also our ensemble results, for a better comparison with these baselines. In the tables, we indicate ensemble methods postponing "(ens.)" to the method name.
}

\majorrevised{We used the original implementations from their respective GitHub repositories to compute the NDCG metrics for the baselines, where possible.}
In the case of missing pre-trained models, we were not able to produce consistent results with the original papers. In this case, we do not report the NDCG metrics ("-").



\setlength{\tabcolsep}{4pt}
\newcolumntype{C}{>{\centering\arraybackslash}p{0.2cm}}
\newcolumntype{R}{D{,}{\pm}{1.2}}
\newcolumntype{L}{>{\raggedright\arraybackslash}p{3.2cm}}
\begin{table}[t]
\centering
\begin{threeparttable}
\caption{Results on the MS-COCO dataset, on the 1K test set.}
\begin{tabular}{LCCCCCCCCCC}
\toprule
& \multicolumn{5}{c}{\textbf{Image Retrieval}} & \multicolumn{5}{c}{\textbf{Sentence Retrieval}} \\
\cmidrule(lr){2-6} \cmidrule(lr){7-11}
& \multicolumn{3}{c}{Recall@K} & \multicolumn{2}{c}{NDCG} & \multicolumn{3}{c}{Recall@K} & \multicolumn{2}{c}{NDCG} \\
\cmidrule(lr){2-4} \cmidrule(lr){5-6} \cmidrule(lr){7-9} \cmidrule(lr){10-11}
\textbf{Model} & \multicolumn{1}{c}{K=1} & \multicolumn{1}{c}{K=5} & \multicolumn{1}{c}{K=10}
& \multicolumn{1}{c}{\texttt{\small{ROUGE-L}}} & \multicolumn{1}{c}{\texttt{\small{SPICE}}} & \multicolumn{1}{c}{K=1} & \multicolumn{1}{c}{K=5} & \multicolumn{1}{c}{K=10}
& \multicolumn{1}{c}{\texttt{\small{ROUGE-L}}} & \multicolumn{1}{c}{\texttt{\small{SPICE}}} \\
\midrule
\multicolumn{11}{c}{(VGG)} \\
\majorrevised{JGCAR \cite{wang2018joint}} & \majorrevised{40.2} & \majorrevised{74.8} & \majorrevised{85.7} & - & - & \majorrevised{52.7} & \majorrevised{82.6} & \majorrevised{90.5} & - & - \\
\majorrevised{SAN \cite{ji2019saliency}} & \majorrevised{60.8} & \majorrevised{90.3} & \majorrevised{95.7} & - & - & \majorrevised{74.9} & \majorrevised{94.9} & \majorrevised{98.2} & - & - \\
\midrule
\multicolumn{11}{c}{(ResNet)} \\
VSE++ \cite{vsepp2018faghri} \tnote{\textdagger} & 52.0 & 84.3 & 92.0 & 0.712 & 0.617 & 64.6 & 90.0 & 95.7 & 0.705	 & 0.658 \\
\majorrevised{SMAN \cite{ji2020sman}} & \majorrevised{58.8} & \majorrevised{87.4} & \majorrevised{93.5} & - & - & \majorrevised{68.4} & \majorrevised{91.3} & \majorrevised{96.6} & - & - \\
\majorrevised{M3A-Net \cite{ji2020multi}} & \majorrevised{58.4} & \majorrevised{87.1} & \majorrevised{94.0} & - & - & \majorrevised{70.4} & \majorrevised{91.7} & \majorrevised{96.8} & - & - \\
\majorrevised{AAMEL \cite{wei2020adversarial}} & \majorrevised{59.9} & \majorrevised{89.0} & \majorrevised{95.1} & - & - & \majorrevised{74.3} & \majorrevised{95.4} & \majorrevised{98.2} & - & - \\
\midrule
\multicolumn{11}{c}{(Region CNN)} \\
MRNN \cite{karpathy2015alignment} & 27.4 & 60.2 & 74.8 & - & - & 38.4 & 69.9 & 80.5 & - & - \\
SCAN (ens.) \cite{lee2018stackedcrossattention} & 58.8 & 88.4 & 94.8 & - & - & 72.7 & 94.8 & 98.4 & -	& - \\
\majorrevised{SAEM (ens.)\cite{wu2019learning} \tnote{\S} \tnote{\textdagger}} & \majorrevised{57.8} & \majorrevised{88.6} & \majorrevised{94.9} & - & - & \majorrevised{71.2} & \majorrevised{94.1} & \majorrevised{97.7} & - & - \\
\majorrevised{CASC \cite{xu2020cross}} & \majorrevised{58.9} & \majorrevised{89.8} & \majorrevised{96.0} & - & - & \majorrevised{72.3} & \majorrevised{96.0} & \majorrevised{\textbf{99.0}} & - & - \\
\majorrevised{MMCA \cite{wei2020multi} \tnote{\S}} & \majorrevised{61.6} & \majorrevised{89.8} & \majorrevised{95.2} & - & - & \majorrevised{74.8} & \majorrevised{95.6} & \majorrevised{97.7} & - & - \\
VSRN \cite{li2019} \tnote{\textdagger}& 60.8 & 88.4 & 94.1 & 0.723 & 0.621 & 74.0 & 94.3 & 97.8 & 0.737 & 0.690 \\
VSRN (ens.) \cite{li2019} \tnote{\textdagger} &  62.8 & 89.7 & 95.1 & 0.732 & 0.637 & 76.2 & 94.8 & 98.2 & 0.748	& 0.704 \\
\majorrevised{PFAN (ens.) \cite{wang2019position}} & \majorrevised{61.6} & \majorrevised{89.6} & \majorrevised{95.2} & - & - & \majorrevised{76.5} & \majorrevised{96.3} & \majorrevised{\textbf{99.0}} & - & - \\
Full-IMRAM \cite{Chen2020imram} &  61.7	& 89.1 & 95.0 & - & - & 76.7 & 95.6 & 98.5 & - & - \\	
\majorrevised{CAMERA \cite{qu2020context} \tnote{\S} \tnote{\textdagger}} & \majorrevised{62.3} & \majorrevised{90.1} & \majorrevised{95.2} & - & - & \majorrevised{75.9} & \majorrevised{95.5} & \majorrevised{98.6} & - & -\\ \majorrevised{CAMERA (ens.) \cite{qu2020context} \tnote{\S} \tnote{\textdagger}} & \majorrevised{63.4} & \majorrevised{90.9} & \majorrevised{95.8} & - & - & \majorrevised{77.5} & \majorrevised{96.3} & \majorrevised{98.8} & - & - \\
\midrule
TERN \cite{messina2020tern} & 51.9 & 85.6 & 93.6 & 0.725 & 0.653 & 63.7 & 90.5 & 96.2 & 0.716 & 0.674 \\
TERN $M_{r}S_{w}$Test & 51.5 & 84.9 & 93.1 & 0.722 & 0.642 & 26.6 & 70.3 & 86.3 &  0.568 & 0.530\\
TERN $M_{w}S_{r}$Test & 51.2 & 84.6 & 92.9 & 0.722 & 0.643 & 61.9 & 88.9 & 95.7 & 0.713 & 0.666 \\
TERN w. Align & 54.5 & 86.9 & 94.2 & 0.724 & 0.643 & 65.5 & 91.0 & 96.5 & 0.720 & 0.675\\
\midrule
TERAN Symm. & 63.5 & 91.1 & 96.3 & 0.739 & 0.666 & 76.3 & 95.3 & 98.4 & 0.741 & 0.701 \\
TERAN $M_{w}S_{r}$ & 57.5 & 88.4 & 94.9 & 0.730 & 0.658 & 70.8 & 93.5 & 97.3 & 0.725 & 0.681 \\
TERAN $M_{r}S_{w}$ & 65.0 & 91.2 & 96.4 & 0.741 & 0.668	& 77.7 & 95.9 & 98.6 & 0.746 & 0.707 \\
\majorrevised{TERAN $M_{r}S_{w}$ (ens.)} & \majorrevised{\textbf{67.0}} & \majorrevised{\textbf{92.2}} & \majorrevised{\textbf{96.9}} & \majorrevised{\textbf{0.747}} & \majorrevised{\textbf{0.680}} & \majorrevised{\textbf{80.2}} & \majorrevised{\textbf{96.6}} & \majorrevised{\textbf{99.0}} & \majorrevised{\textbf{0.756}} & \majorrevised{\textbf{0.720}} \\
\bottomrule
\end{tabular}
\label{tab:results_mscoco_1k}
\begin{tablenotes}
    \item[\S] Uses BERT as language model
    \item[\textdagger] Uses disentangled visual-textual pipelines
\end{tablenotes}
\end{threeparttable}
\end{table}

\setlength{\tabcolsep}{4pt}
\newcolumntype{C}{>{\centering\arraybackslash}p{0.2cm}}
\newcolumntype{R}{D{,}{\pm}{1.2}}
\newcolumntype{L}{>{\raggedright\arraybackslash}p{3.2cm}}
\begin{table}[t]
\centering
\begin{threeparttable}
\caption{Results on the MS-COCO dataset, on the 5K test set.}
\begin{tabular}{LCCCCCCCCCC}
\toprule
& \multicolumn{5}{c}{\textbf{Image Retrieval}} & \multicolumn{5}{c}{\textbf{Sentence Retrieval}} \\
\cmidrule(lr){2-6} \cmidrule(lr){7-11}
& \multicolumn{3}{c}{Recall@K} & \multicolumn{2}{c}{NDCG} & \multicolumn{3}{c}{Recall@K} & \multicolumn{2}{c}{NDCG} \\
\cmidrule(lr){2-4} \cmidrule(lr){5-6} \cmidrule(lr){7-9} \cmidrule(lr){10-11}
\textbf{Model} & \multicolumn{1}{c}{K=1} & \multicolumn{1}{c}{K=5} & \multicolumn{1}{c}{K=10}
& \multicolumn{1}{c}{\texttt{\small{ROUGE-L}}} & \multicolumn{1}{c}{\texttt{\small{SPICE}}} & \multicolumn{1}{c}{K=1} & \multicolumn{1}{c}{K=5} & \multicolumn{1}{c}{K=10}
& \multicolumn{1}{c}{\texttt{\small{ROUGE-L}}} & \multicolumn{1}{c}{\texttt{\small{SPICE}}} \\
\midrule
\multicolumn{11}{c}{(ResNet)} \\
VSE++ \cite{vsepp2018faghri} \tnote{\textdagger}& 30.3 & 59.4 & 72.4 & 0.656 & 0.577 & 41.3 & 71.1 & 81.2 & 0.597 & 0.551 \\
\majorrevised{M3A-Net \cite{ji2020multi}} & \majorrevised{38.3} & \majorrevised{65.7} & \majorrevised{76.9} & - & - & \majorrevised{48.9} & \majorrevised{75.2} & \majorrevised{84.4} & - & - \\
\majorrevised{AAMEL \cite{wei2020adversarial}} & \majorrevised{39.9} & \majorrevised{71.3} & \majorrevised{81.7} & - & - & \majorrevised{51.9} & \majorrevised{84.2} & \majorrevised{91.2} & - & - \\
\midrule
\multicolumn{11}{c}{(Region CNN)} \\
MRNN \cite{karpathy2015alignment} & 10.7 & 29.6 & 42.2 & - & - & 16.5 & 39.2 & 52.0 & - & -\\
SCAN (ens.) \cite{lee2018stackedcrossattention} &  38.6 & 69.3 & 80.4 & - & - & 50.4 & 82.2 & 90.0 & -	& - \\
VSRN \cite{li2019} \tnote{\textdagger} & 37.9 & 68.5 & 79.4 & 0.676 & 0.596 & 50.3 & 79.6 & 87.9 & 0.639 & 0.598 \\
VSRN (ens.) \cite{li2019} \tnote{\textdagger}&  40.5	& 70.6 & 81.1 & 0.684 & 0.609 & 53.0 & 81.1 & 89.4 & 0.652 & 0.612 \\
Full-IMRAM \cite{Chen2020imram} &  39.7 & 69.1 & 79.8 & - & - & 53.7 & 83.2 & 91.0 & - & - \\
\majorrevised{MMCA \cite{wei2020multi} \tnote{\S}} & \majorrevised{38.7} & \majorrevised{69.7} & \majorrevised{80.8} & - & - & \majorrevised{54.0} & \majorrevised{82.5} & \majorrevised{90.7} & - & - \\
\majorrevised{CAMERA \cite{qu2020context} \tnote{\S} \tnote{\textdagger}} & \majorrevised{39.0} & \majorrevised{70.5} & \majorrevised{81.5} & - & - & \majorrevised{53.1} & \majorrevised{81.3} & \majorrevised{89.8} & - & - \\
\majorrevised{CAMERA (ens.) \cite{qu2020context} \tnote{\S} \tnote{\textdagger}} & \majorrevised{40.5} & \majorrevised{71.7} & \majorrevised{82.5} & - & - & \majorrevised{55.1} & \majorrevised{82.9} & \majorrevised{91.2} & - & - \\

\midrule
TERN \cite{messina2020tern} & 28.7 & 59.7 & 72.7 & 0.665 & 0.599 & 38.4 & 69.5 & 81.3 & 0.601 & 0.556\\
TERN $M_{r}S_{w}$Test & 28.3 & 59.1 & 72.2 & 0.663 & 0.592 & 6.8 & 28.4 & 46.7 & 0.406 & 0.372 \\
TERN $M_{w}S_{r}$Test & 28.1 & 58.6 & 71.8 & 0.663 & 0.592 & 35.5 & 67.5 & 78.9 & 0.600 & 0.551 \\
TERN w. Align & 31.4 & 62.5 & 75.3 & 0.667 & 0.597 & 40.2 & 71.1 & 81.9 & 0.606 & 0.561 \\
\midrule
TERAN Symm. & 41.0 & 71.6 & 82.3 & 0.680 & 0.607 & 54.8 & 82.7 & 90.9 & 0.641 & 0.601\\
TERAN $M_{w}S_{r}$ & 34.1 & 65.7 & 77.8 & 0.669 & 0.596 & 45.3 & 76.3 & 86.2 & 0.611 & 0.564\\
TERAN $M_{r}S_{w}$ & 42.6 & 72.5 & 82.9 & 0.682 & 0.610 & 55.6 & 83.9 & 91.6 & 0.643 & 0.606 \\
\majorrevised{TERAN $M_{r}S_{w}$(ens.)} & \majorrevised{\textbf{45.1}} & \majorrevised{\textbf{74.6}} & \majorrevised{\textbf{84.4}} & \majorrevised{\textbf{0.689}} & \majorrevised{\textbf{0.622}} & \majorrevised{\textbf{59.3}} & \majorrevised{\textbf{85.8}} & \majorrevised{\textbf{92.4}} & \majorrevised{\textbf{0.658}} & \majorrevised{\textbf{0.624}} \\
\bottomrule
\end{tabular}
\label{tab:results_mscoco_5k}
\begin{tablenotes}
    \item[\S] Uses BERT as language model
    \item[\textdagger] Uses disentangled visual-textual pipelines
\end{tablenotes}
\end{threeparttable}
\end{table}



\majorrevised{
On both the 1K and 5K test sets, our novel TERAN approach reaches state-of-the-art results on almost all the metrics.
Concerning the results reported in Table \ref{tab:results_mscoco_1k} regarding 1K test set, the best performing TERAN model is the one implementing the max-over-regions sum-over-words ($M_{r}S_{w}$) pooling method, although the model using the symmetric loss reaches comparable results. We chose the same TERAN $M_{r}S_{w}$ model to evaluate the ensemble, reaching an improvement of 5.7\% and 3.5\% on the Recall@1 metric on image and sentence retrieval respectively, with respect to the best baseline using ensemble methods, which is CAMERA \cite{qu2020context}. Notice, however, that even the basic TERAN model without ensemble is able to surpass CAMERA in many metrics. This confirms the power of the TERAN model despite its overall simplicity.

Table \ref{tab:results_mscoco_5k} reports the results for the 5K test set, which confirm the superiority of TERAN $M_{r}S_{w}$ over all the baselines also on the full test set. In this scenario, we increase the Recall@1 performance by 11.3\% and 7.6\% on image and sentence retrieval with respect to the CAMERA approach.}
On the other hand, the max-over-words sum-over-regions ($M_{w}S_{r}$) method loses around 10\% on the Recall@1 metrics with respect to the best performing TERAN non-ensemble model. In this case, the Recall@K metric does not improve over top results obtained by the current state-of-the-art approaches. Nevertheless, this model loses only about 1.5\% during image-retrieval and about 3.5\% during sentence-retrieval as far as the \texttt{SPICE} NDCG metric is concerned, reaching perfectly comparable results with our state-of-the-art method.
In light of these results, we deduce that the $M_{w}S_{r}$ model is not so effective in retrieving the perfect-matching elements; however, it is still very good at retrieving the relevant ones.

As far as image retrieval is concerned, in the \textit{TERN $M_{w}S_{r}$Test} and \textit{TERN $M_{w}S_{r}$Test} experiments we can see that the TERN architecture trained as in \cite{messina2020tern} performs fairly good when the similarity is computed as in the novel TERAN architecture, using the region and words outputs and not the I-CLS and T-CLS global descriptions. In particular, the use of max-over-words sum-over-regions similarity still works quite well compared to the similarity computed through I-CLS and T-CLS global visual and textual features as it is in TERN.

Notice instead that on the sentence retrieval task, the \textit{TERN $M_{r}S_{w}$ Test} experiment obtains a very low performance. This is the consequence of the fact that TERN is trained to produce global-scale image-sentence matchings, while it is never forced to produce meaningful fine-grained aligned concepts. This is further supported by the evidence that if we visualize the region-words alignments as explained in the following Section \ref{sec:visualize_alignments} we obtain random word groundings on the image, meaning that the concepts in output from TERN are not sufficiently informative.

\begin{figure}[t]
\begin{subfigure}[t]{0.49\textwidth}
\centering
\includegraphics[width=\linewidth]{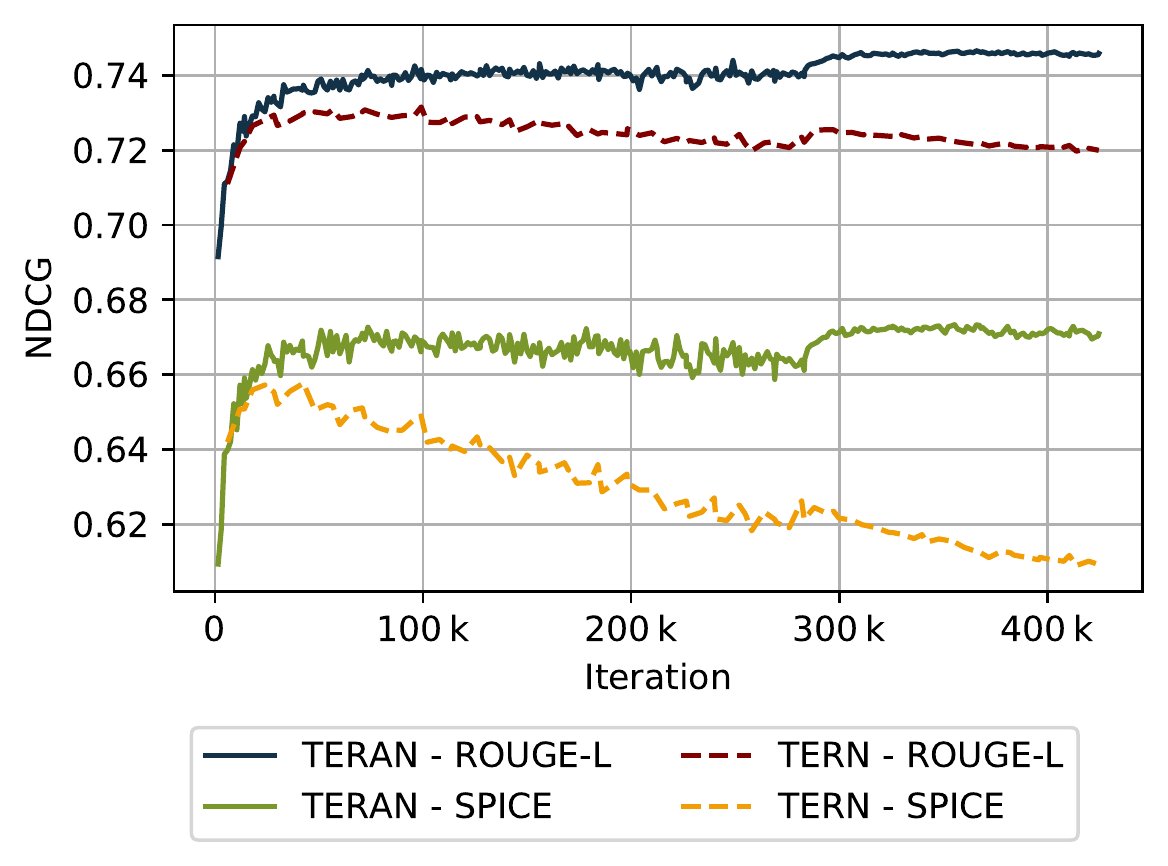}
\end{subfigure}
\begin{subfigure}[t]{0.473\textwidth}
\centering
\includegraphics[width=\linewidth]{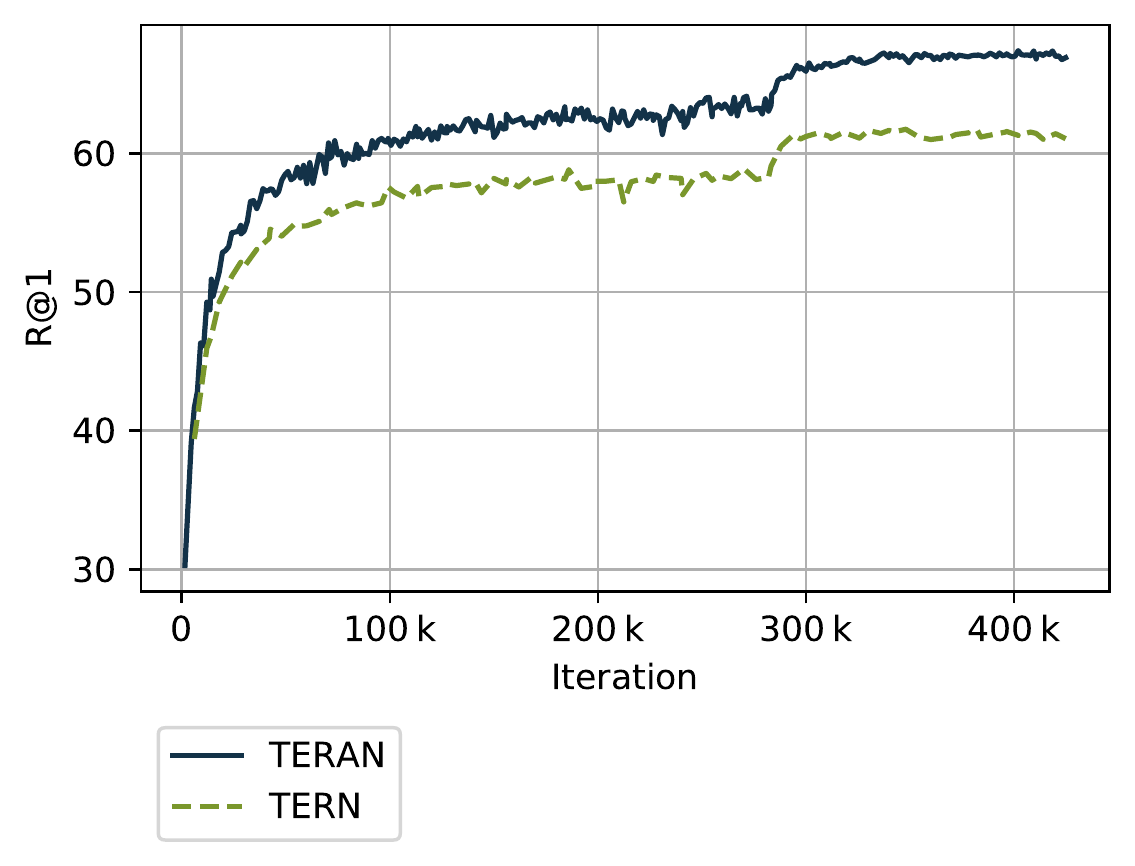}
\end{subfigure}
\caption{\majorrevised{Validation metrics on sentence-to-image retrieval, measured during the training phase, for the average-over-sentences scenario. TERN overfits on the NDCG metrics, while Recall@1 still improves. TERAN instead generalizes better on both metrics.}}
\label{fig:tern_overfit}       
\end{figure}

\majorrevised{In order to better compare TERAN with our previous TERN approach, in Figure \ref{fig:tern_overfit} we report the validation curves for both NDCG and Recall@1 metrics, for both methods. We can notice how the NDCG metric overfits in our previous TERN model, especially when using the SPICE metric, while the Recall@ keeps increasing. On the other hand, TERAN demonstrates better generalization abilities on both metrics. This is a clear indication that TERAN is better able to retrieve relevant items in the first positions, as well as exact matching elements. Instead, TERN is more prone to overfitting to the SPICE metric, meaning that at a certain point in training, the network still searches for the top matching element, but with a tendency to push away possible relevant results compared to the novel TERAN approach.}

\setlength{\tabcolsep}{4pt}
\newcolumntype{C}{>{\centering\arraybackslash}p{0.15cm}}
\newcolumntype{R}{D{,}{\pm}{1.2}}
\newcolumntype{L}{>{\raggedright\arraybackslash}p{3.2cm}}
\begin{table}[t]
\centering
\begin{threeparttable}
\caption{Results on the Flickr30k dataset.}
\begin{tabular}{LCCCCCCCCCC}
\toprule
& \multicolumn{5}{c}{\textbf{Image Retrieval}} & \multicolumn{5}{c}{\textbf{Sentence Retrieval}} \\
\cmidrule(lr){2-6} \cmidrule(lr){7-11}
& \multicolumn{3}{c}{Recall@K} & \multicolumn{2}{c}{NDCG} & \multicolumn{3}{c}{Recall@K} & \multicolumn{2}{c}{NDCG} \\
\cmidrule(lr){2-4} \cmidrule(lr){5-6} \cmidrule(lr){7-9} \cmidrule(lr){10-11}
\textbf{Model} & \multicolumn{1}{c}{K=1} & \multicolumn{1}{c}{K=5} & \multicolumn{1}{c}{K=10}
& \multicolumn{1}{c}{\texttt{\small{ROUGE-L}}} & \multicolumn{1}{c}{\texttt{\small{SPICE}}} & \multicolumn{1}{c}{K=1} & \multicolumn{1}{c}{K=5} & \multicolumn{1}{c}{K=10}
& \multicolumn{1}{c}{\texttt{\small{ROUGE-L}}} & \multicolumn{1}{c}{\texttt{\small{SPICE}}} \\
\midrule
\multicolumn{11}{c}{(VGG)} \\
\majorrevised{JGCAR \cite{wang2018joint}} & \majorrevised{35.2} & \majorrevised{62.0} & \majorrevised{72.4} & - & - & \majorrevised{44.9} & \majorrevised{75.3} & \majorrevised{82.7} & - & - \\
\majorrevised{SAN \cite{ji2019saliency}} & \majorrevised{51.4} & \majorrevised{77.2} & \majorrevised{85.2} & - & - & \majorrevised{67.0} & \majorrevised{88.0} & \majorrevised{94.6} & - & -\\ 
\midrule
\multicolumn{11}{c}{(ResNet)} \\
VSE++ \cite{vsepp2018faghri} \tnote{\textdagger}& 39.6	& 70.1 & 79.5 & 0.631 & 0.494 & 52.9 & 80.5 & 87.2 & 0.601 & 0.514 \\
\majorrevised{TIMAM \cite{sarafianos2019adversarial} \tnote{\S} \tnote{\textdagger}} & \majorrevised{42.6} & \majorrevised{71.6} & \majorrevised{81.9} & - & - & \majorrevised{53.1} & \majorrevised{78.8} & \majorrevised{87.6} & - & -\\
\majorrevised{SMAN \cite{ji2020sman}} & \majorrevised{43.4} & \majorrevised{73.7} & \majorrevised{83.4} & - & - & \majorrevised{57.3} & \majorrevised{85.3} & \majorrevised{92.2} & - & - \\
\majorrevised{M3A-Net \cite{ji2020multi}} & \majorrevised{44.7} & \majorrevised{72.4} & \majorrevised{81.1} & - & - & \majorrevised{58.1} & \majorrevised{82.8} & \majorrevised{90.1} & - & - \\
\majorrevised{AAMEL \cite{wei2020adversarial}} & \majorrevised{49.7} & \majorrevised{79.2} & \majorrevised{86.4} & - & - & \majorrevised{68.5} & \majorrevised{91.2} & \majorrevised{95.9} & - & - \\
\midrule
\multicolumn{11}{c}{(Region CNN)} \\
MRNN \cite{karpathy2015alignment} & 15.2 & 37.7 & 50.5 & - & - & 22.2 & 48.2 & 61.4 & - & - \\
SCAN (ens.) \cite{lee2018stackedcrossattention} &  48.6 & 77.7 & 85.2 & - & - & 67.4 & 90.3 & 95.8 & -	& - \\
\majorrevised{PFAN (ens.) \cite{wang2019position}} & \majorrevised{50.4} & \majorrevised{78.7} & \majorrevised{86.1} & - & - & \majorrevised{70.0} & \majorrevised{91.8} & \majorrevised{95.0} & - & - \\
\majorrevised{SAEM (ens.)\cite{wu2019learning} \tnote{\S} \tnote{\textdagger}} & \majorrevised{52.4} & \majorrevised{81.1} & \majorrevised{88.1} & - & - & \majorrevised{69.1} & \majorrevised{91.0} & \majorrevised{95.1} & - & - \\
VSRN \cite{li2019} \tnote{\textdagger} & 53.0 & 77.9 & 85.7 & 0.673 & 0.545 & 70.4 & 89.2 & 93.7 & 0.676 & 0.592 \\
VSRN (ens.) \cite{li2019} \tnote{\textdagger} &  54.7 & 81.8 & 88.2 & 0.680 & 0.556 & 71.3 & 90.6 & 96.0 & 0.688 & 0.606 \\
Full-IMRAM \cite{Chen2020imram} &  53.9 & 79.4 & 87.2 & - & - & 74.1 & 93.0 & 96.6 & - & - \\	
\majorrevised{MMCA \cite{wei2020multi} \tnote{\S}} & 54.8 & 81.4 & 87.8 & - & - & 74.2 & 92.8 & 96.4 & - & - \\
\majorrevised{CASC \cite{xu2020cross}} & \majorrevised{60.2} & \majorrevised{78.3} & \majorrevised{86.3} & - & - & \majorrevised{68.5} & \majorrevised{90.6} & \majorrevised{95.9} & - & - \\
\majorrevised{CAMERA \cite{qu2020context} \tnote{\S} \tnote{\textdagger}} & \majorrevised{58.9} & \majorrevised{84.7} & \majorrevised{90.2} & - & - & \majorrevised{76.5} & \majorrevised{\textbf{95.1}} & \majorrevised{97.2} & - & - \\
\majorrevised{CAMERA (ens.) \cite{qu2020context} \tnote{\S} \tnote{\textdagger}} & \majorrevised{60.3} & \majorrevised{85.9} & \majorrevised{91.7} & - & - & \majorrevised{78.0} & \majorrevised{\textbf{95.1}} & \majorrevised{\textbf{97.9}} & - & -\\
TERN \cite{messina2020tern} & 41.1 & 71.9 & 81.2 & 0.647 & 0.512 & 53.2 & 79.4 & 86.0 & 0.624  & 0.529 \\
\midrule
TERAN Symm. & 55.7 & 83.1 & 89.3 & 0.678 & 0.555 & 71.8 & 90.5 & 94.7 & 0.676 & 0.603 \\
TERAN $M_{w}S_{r}$ & 49.4 & 78.3 & 85.9 & 0.664 & 0.536 & 60.5 & 85.1 & 92.2 & 0.651 & 0.558 \\
TERAN $M_{r}S_{w}$ & 59.5 & 84.9 & 90.6 & 0.686 & 0.564 & 75.8 & 93.2 & 96.7 & 0.687 &  0.614\\
\majorrevised{TERAN $M_{r}S_{w}$ (ens.)} & \majorrevised{\textbf{63.1}} & \majorrevised{\textbf{87.3}} & \majorrevised{\textbf{92.6}} & \majorrevised{\textbf{0.695}} & \majorrevised{\textbf{0.577}} & \majorrevised{\textbf{79.2}} & \majorrevised{94.4} & \majorrevised{96.8} & \majorrevised{\textbf{0.707}} & \majorrevised{\textbf{0.636}} \\
\bottomrule
\end{tabular}
\label{tab:results_f30k}
\begin{tablenotes}
    \item[\S] Uses BERT as language model
    \item[\textdagger] Uses disentangled visual-textual pipelines
\end{tablenotes}
\end{threeparttable}
\end{table}


\majorrevised{
However, looking at the results from the \textit{TERN w. Align} experiment, we can notice that by augmenting the TERN objective with the TERAN alignment loss, we can slightly increase the TERN overall performance. This confirms that a more precise and meaningful region-word alignment has a visible effect also on the quality of the fixed-sized global embeddings produced by TERN.}

\majorrevised{In Table \ref{tab:results_f30k} we report the results on the Flickr30k dataset. Our single-model TERAN $M_{r}S_{w}$ method outperforms the best baseline (CAMERA) on the image retrieval task while approaching the single-model CAMERA performance on the sentence retrieval task. Nevertheless, even on Flickr30k our TERAN $M_{r}S_{w}$ method with model ensemble obtains state-of-the-art results with respect to all the baselines on all the metrics, gaining 4.6\% and 1.5\% on the Recall@1 metric on the image and sentence retrieval tasks respectively.} 

\revised{On the MS-COCO dataset, our system powered by a single GTX 1080Ti can compute a single image-to-sentence query in $\sim0.12 s$ on 5k sentences of the test split; in the sentence-to-image scenario, it can produce scores and rank the 1K images in $\sim0.02 s$. These timings allow the TERAN scores to be effectively used, for example, in a re-ranking phase, where the first 1k images - 5k sentences have been previously retrieved using a faster descriptor (e.g., the one from TERN).}

\subsection{Qualitative Analysis for Image Retrieval}
\begin{figure*}[t]
    \centering
    \includegraphics[page=2, width=1\linewidth]{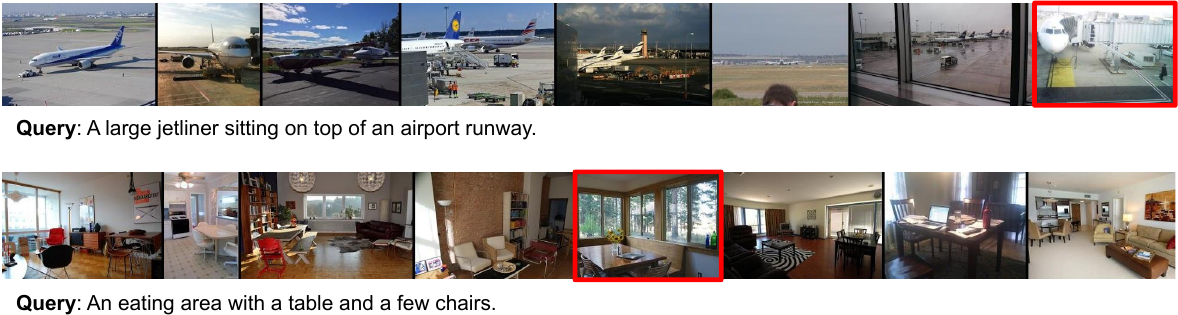}
  \caption{Example of image retrieval results for a couple of flexible query captions. These are common examples where NDCG succeeds over the Recall@K metric. The ground-truth matching image is not among the very first positions; however, the top-ranked images are also visually very relevant.}
  \label{fig:examples_ndcg} 
\end{figure*}

\begin{figure*}[t]
    \centering
    \includegraphics[page=5, width=1\linewidth]{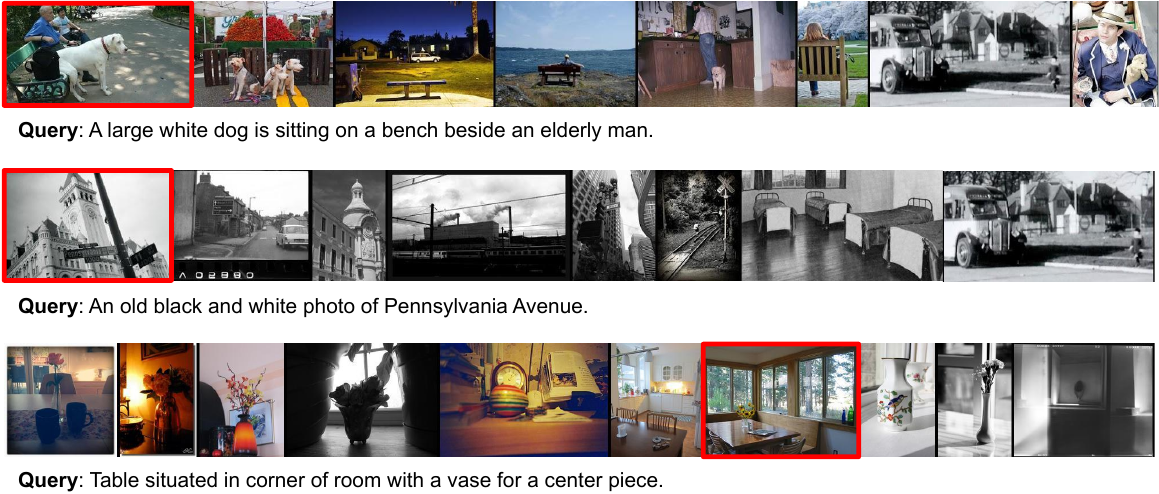}
  \caption{Example of image retrieval results for a couple of very specific query captions.}
  \label{fig:examples_exact_matching} 
\end{figure*}

The visualization of image retrieval results is a good way to qualitatively appreciate the retrieval abilities of the proposed TERAN model.
Figures \ref{fig:examples_ndcg} and \ref{fig:examples_exact_matching} show examples of images retrieved given a textual caption as a query, with scores computed using the max-over-regions sum-over-words method. 
In particular, Figure \ref{fig:examples_ndcg} shows image retrieval results for a couple of flexible query captions. The red-marked images represent the exact-matching elements from the ground-truth. We can therefore conclude that the retrieved images in these examples are incorrect results for the Recall@1 metric (and for the first query even for Recall@5). However, in the very first positions, we find non-matching yet relevant images, due to the ambiguity of the query caption. These are common examples where NDCG succeeds over the Recall@K metric since we need a flexible evaluation for not-too-strict query captions.

Figure \ref{fig:examples_exact_matching} reports instead image retrieval results for a couple of very specific query captions. For the first two queries, the network succeeds in positioning the only really relevant image in the first position (a dog sitting on a bench on the upper query, and Pennsylvania Avenue, uniquely identifiable by the street sign, on the lower query). In this case, the Recall@1 metric also succeeds, given that the query captions are very selective. \revised{The third example, instead, evidences a failure case where the model cannot deal with very subtle details. The (only) correct result is ranked 6th in this case; in the first ranking positions, the model can find images with a vase used as a centerpiece, but the table is not often visible, and when it is visible, it is not in the corner of the room.}

\section{Ablation Study}
\subsection{The Effect of Weight Sharing}
\label{sec:weight_sharing}
We tried to apply weight sharing for the last 2 layers of the TERAN architecture, those after the linear projection to the 1024-d space. Weight sharing is used to reduce the size of the network and enforce a structure able to perform common reasoning on the high-level concepts, possibly reducing the overfitting and increasing the stability of the whole network.
We experimented with the effects of weight sharing on the MS-COCO dataset with 1K test set, for both the max-over-words sum-over-regions and the max-over-regions sum-over-words scenarios. 

Results are shown in the 2-nd and 6-th rows of Table \ref{tab:secondary_results}. It can be noticed that the values are perfectly comparable with the TERAN results reported in Table \ref{tab:results_mscoco_1k}, suggesting that at this point in the network the abstraction is high enough that concepts coming from images and sentences can be processed in the exact same way. This result shows that vectors at this stage have been freed from any modality bias and they are fully comparable in the same representation space.

Also, in the max-over-words sum-over-regions scenario (6-th row), there is a small gain both in terms of Recall@K and NDCG. This confirms the slight regularization effect of the weight sharing approach.

\subsection{Averaging Versus Summing}
We tried to do average instead of sum during the last pooling phase of the alignment matrix.
We consider only the case in which we average-over-sentences; in fact, since in our experiments the number of visual concepts is always fixed to the 36 more influent ones during the object detection stage, average-over-regions and sum-over-regions do not differ substantially.

Thus, we considered the case of max-over-regions average-over-words ($M_{r}\text{Avg}_{w}$):
\begin{equation}
    S_{kl} = \frac{\sum_{j \in g_l}\max_{i \in g_k} {A_{ij}}}{|g_l|}.
\end{equation}

If we compute the average instead of the sum in the max-over-regions sum-over-words scenario, the final similarity score between the image and the sentence is no more dependent on the number of concepts from the textual pipeline: the similarities are averaged and not accumulated.

In the 3-rd row of Table \ref{tab:secondary_results} we can notice that by averaging we lose an important amount of information with respect to the max-over-regions sum-over-words scenario (1-st row). This insight suggests that the complexity of the query is beneficial for achieving high-quality matching. 

\begin{figure}[t]
\begin{subfigure}[b]{0.49\textwidth}
\centering
\includegraphics[width=\linewidth]{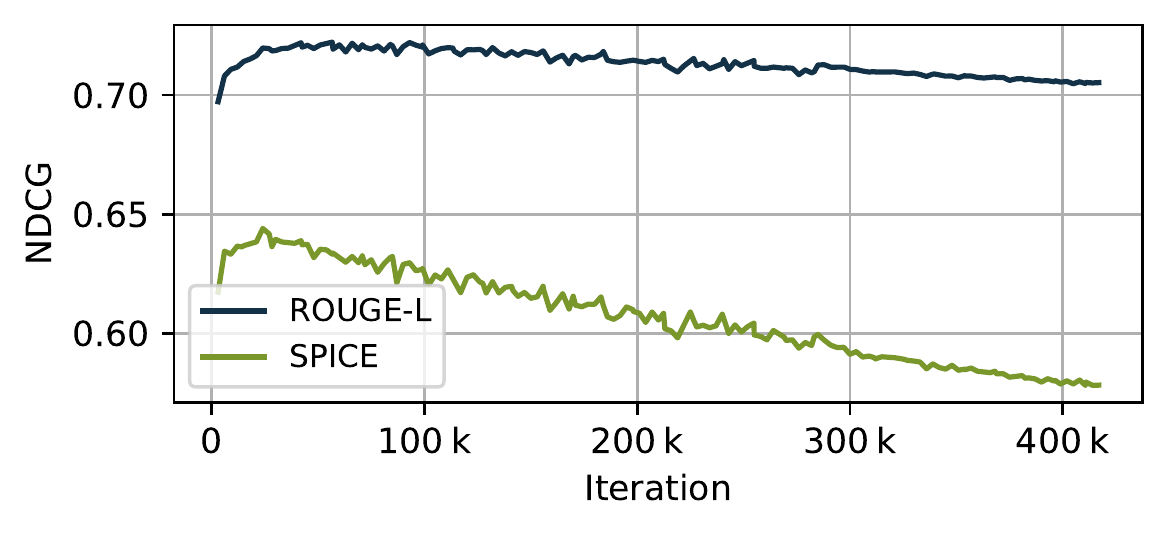}
\end{subfigure}
\begin{subfigure}[b]{0.476\textwidth}
\centering
\includegraphics[width=\linewidth]{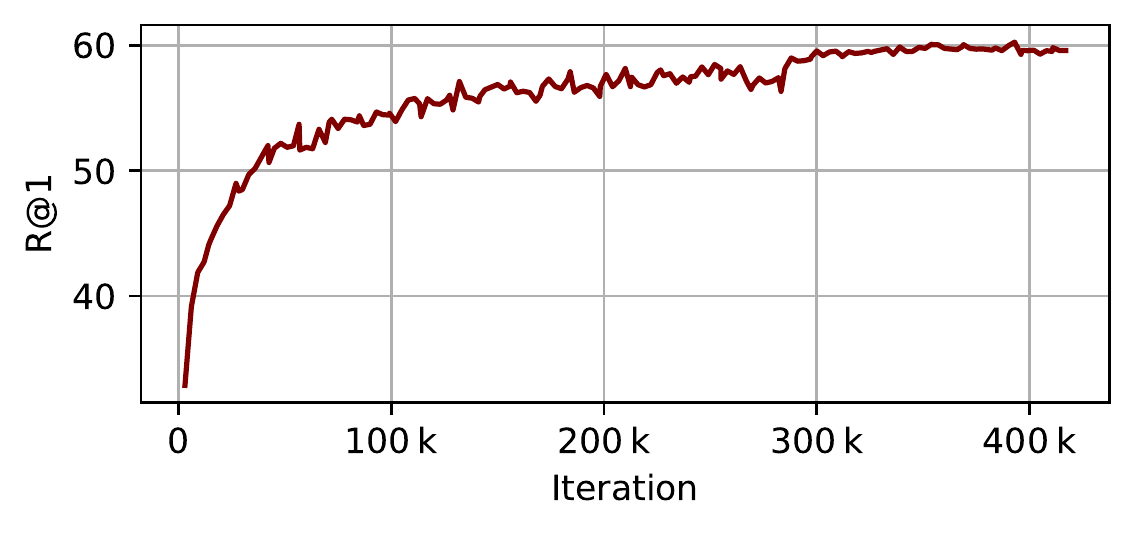}
\end{subfigure}
\caption{Validation metrics measured during the training phase, for the  average-over-sentences scenario. This model overfits on the NDCG metrics on the image-retrieval task, while Recall@1 still improves.}
\label{fig:overfit}       
\end{figure}

Another side effect of using average instead of the max is the premature clear overfitting on the NDCG metrics as far as image-retrieval is concerned. The effect is shown in Figure \ref{fig:overfit}. \majorrevised{The clear overfitting of the NDCG metrics resembles the training curve trajectories of TERN (Figure \ref{fig:tern_overfit})}. This result demonstrates that although this model can correctly perform exact matching, it is pulling away relevant results from the head of the ranked list of images, during the validation phase. 

\setlength{\tabcolsep}{4pt}
\newcolumntype{C}{>{\centering\arraybackslash}p{0.15cm}}
\newcolumntype{R}{D{,}{\pm}{1.2}}
\newcolumntype{L}{>{\raggedright\arraybackslash}p{3cm}}
\begin{table}[t]
\begin{center}
\caption{Results for the ablation study experiments. We organize the methods in the table clustering them by the pooling method, for an easier comparison (max-over-regions methods in the upper part and max-over-words methods on the lower part). In the first row of both sections we report the TERAN results from Table \ref{tab:results_mscoco_1k}. Experiments are computed on the MS-COCO dataset, 1K test set.}
\begin{tabular}{LCCCCCCCCCC}
\toprule
& \multicolumn{5}{c}{\textbf{Image Retrieval}} & \multicolumn{5}{c}{\textbf{Sentence Retrieval}} \\
\cmidrule(lr){2-6} \cmidrule(lr){7-11}
& \multicolumn{3}{c}{Recall@K} & \multicolumn{2}{c}{NDCG} & \multicolumn{3}{c}{Recall@K} & \multicolumn{2}{c}{NDCG} \\
\cmidrule(lr){2-4} \cmidrule(lr){5-6} \cmidrule(lr){7-9} \cmidrule(lr){10-11}
\textbf{Model} & \multicolumn{1}{c}{K=1} & \multicolumn{1}{c}{K=5} & \multicolumn{1}{c}{K=10}
& \multicolumn{1}{c}{\texttt{\small{ROUGE-L}}} & \multicolumn{1}{c}{\texttt{\small{SPICE}}} & \multicolumn{1}{c}{K=1} & \multicolumn{1}{c}{K=5} & \multicolumn{1}{c}{K=10}
& \multicolumn{1}{c}{\texttt{\small{ROUGE-L}}} & \multicolumn{1}{c}{\texttt{\small{SPICE}}} \\
\midrule
$M_{r}S_{w}$ \small{\textit{(from Table \ref{tab:results_mscoco_1k})}} & 65.0 & 91.2 & 96.4 & 0.741	& 0.668	& 77.7 & 95.9 & 98.6 & 0.746 & 0.707 \\
$M_{r}S_{w}$ \small{Shared-W} & 64.5 & 91.3 & 96.3 & 0.740 & 0.667 & 77.3 & 95.9 & 98.4 & 0.746 & 0.706 \\
$M_{r}\text{Avg}_{w}$ & 57.2 & 87.6 & 93.6 & 0.705 & 0.587 & 68.6 & 92.4 & 96.7 & 0.721 & 0.671 \\
$M_{r}S_{w}$ \small{StopWordsFilter} & 64.2 & 91.1 & 96.3 & 0.737 & 0.658 & 76.8 & 95.9 & 98.6 & 0.745 & 0.705\\
\majorrevised{$M_{r}S_{w}$ Bi-LSTM} & \majorrevised{55.6} & \majorrevised{86.9} & \majorrevised{93.9} & \majorrevised{0.734} & \majorrevised{0.666} & \majorrevised{67.4} & \majorrevised{92.5} & \majorrevised{96.9} & \majorrevised{0.717} & \majorrevised{0.677}\\
\majorrevised{$M_{r}S_{w}$ Bi-GRU} & \majorrevised{56.3} & \majorrevised{87.1} & \majorrevised{94.0} & \majorrevised{0.735} & \majorrevised{0.666} & \majorrevised{69.1} & \majorrevised{93.4} & \majorrevised{97.1} & \majorrevised{0.720} & \majorrevised{0.678} \\
\midrule
$M_{w}S_{r}$ \small{\textit{(from Table \ref{tab:results_mscoco_1k})}} & 57.5 & 88.4 & 94.9 & 0.730 & 0.658 & 70.8 & 93.5 & 97.3 & 0.725 & 0.681 \\
$M_{w}S_{r}$ \small{Shared-W}  & 58.1 & 88.4 & 95.0 & 0.730 & 0.657 & 71.1 & 93.1 & 97.7 & 0.728 & 0.683 \\

\bottomrule
\end{tabular}
\label{tab:secondary_results}
\end{center}
\end{table}

\subsection{Removing Stop-Words During Alignment}
Some words may carry no substantial meaning by themselves, such as articles or prepositions. These words with a high and diffuse frequency of use are typically called \textit{stop-words} and are usually removed in classical text analysis processes. In this context, removing stop-words may help the architecture to focus only on the important concepts. Doing so, the training process is simplified as the noise introduced by possibly irrelevant words is removed.
Results are reported in the 4-th row of Table \ref{tab:secondary_results}.
\begin{figure}[t]
    \centering
    \includegraphics[page=6, width=0.9\linewidth]{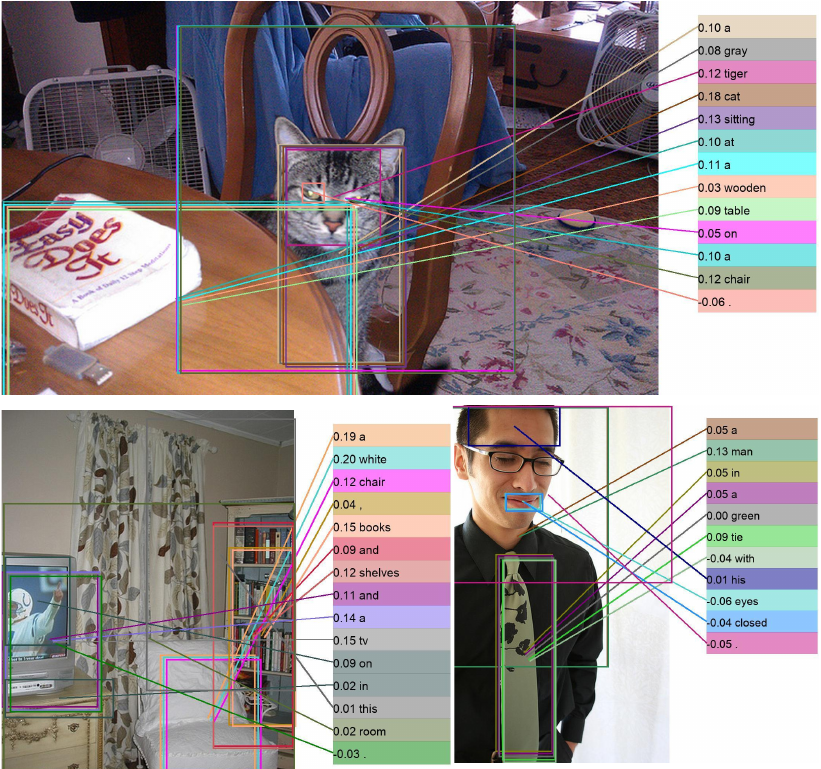}
  \caption{Visualization of the word-region alignments. Near each word, we report the cosine similarity computed between that word and the top-relevant image region associated with it. We slightly offset the overlapping bounding-boxes for a better visualization.}
\label{fig:alignment_visualization}       
\end{figure}
The overall performance, both in terms of Recall@ and NDCG is comparable, yet with a small decrease, with the one obtained without stop-words removal (1-st row of the table). 
This suggests that in this context stop-words are linguistic elements that bring some useful information to distinguish ambiguous scenes. Prepositions and adverbs often indicate the spatial arrangement of objects, thus "chair \textit{near} the table" is not the same as "chair \textit{over} the table"; distinguishing these fine-grained differences is beneficial for obtaining a precise image-text matching.

\majorrevised{
\subsection{Using Different Language Models}
Despite the power of BERT \cite{devlin2019bert} for obtaining contextualized representations for the words in a sentence, many works use recurrent bidirectional networks instead, such as Bi-GRU or Bi-LSTMs. 

In the 5-th and 6-th row of Table \ref{tab:secondary_results} we report the results for the TERAN model with Bi-GRU and Bi-LSTM in substitution of the BERT model for language processing. We used 300-dimensional word embeddings, and a hidden size of 512 so that the final bi-directional sentence feature is a 1024-dimensional description; we used the same training protocol and hyper-parameters used for the main experiments. The results suggest that BERT is an essential ingredient for reaching top results on the Recall@K metrics, especially when K=\{1, 5\}. In particular, Bi-LSTM and Bi-GRU lose around 14\% on image retrieval and 12\% on sentence retrieval on the Recall@1 metric compared to the TERAN $M_{r}S_{w}$ single-model method. However, we can notice that TERAN with these recurrent language models still maintains a comparable performance with respect to the NDCG metric, especially on the image retrieval task.
}

\subsection{Visualizing the Visual-Word Alignments}
\label{sec:visualize_alignments}
Inspired by the work in \cite{karpathy2015alignment}, we try to visualize the region-word alignments learned by TERAN on some images from the test set of MS-COCO dataset. We recall that no supervision was used at the region-word level during the training phase.

In Figure \ref{fig:alignment_visualization}, we report some figures where every sentence word has been associated with the top-relevant image region. The affinity between visual concepts (region features) and textual concepts (word features) has been measured through cosine similarity, just as during the training phase. 

We can see that the words have overall plausible groundings on the image they describe. Some words are really difficult to ground, such as articles, verbs, or adjectives. However, we can notice that phrases describing a visual entity and composed of nouns with the related articles and adjectives (e.g. "a green tie", or "a wooden table") are often grounded to the same region.
This further confirms that the TERAN architecture can produce meaningful concepts, and it is also able to cluster them under the form of complete reasonable phrases.

We can notice some wrong word groundings in the images, such as the phrase "eyes closed" that is associated with the region depicting the closed mouth. In this case, the error seems to lie on some localized misunderstanding of the scene (in this case the noun "eyes" has probably been misunderstood since the mouth and the eyes are both closed).
Overall, however, complex scenes are correctly broken down into their salient elements, and only the key regions are attended.

\section{Conclusions}
In this work, we introduced the Transformer Encoder Reasoning and Alignment Network (TERAN). TERAN is a relationship-aware architecture based on the Transformer Encoder (TE) architecture, exploiting self-attention mechanisms, able to reason about the spatial and abstract relationships between elements in the image and in the text separately. 

Differently from TERN \cite{messina2020tern}, TERAN forces a fine-grained alignment among the region and word features without any supervision at this level.
We demonstrated that by enforcing this fine-grained word-region alignment at training time we can obtain state-of-the-art results on the popular MS-COCO and Flickr30K datasets. \majorrevised{Besides, thanks to the overall simplicity of the proposed model, we can obtain effective visual and textual features for use in scalable retrieval setups.}

We measured the performance of our TERAN architecture in the context of cross-modal retrieval using both the already-in-use Recall@K metric and the newly introduced NDCG with the \texttt{ROUGE-L} and \texttt{SPICE} textual relevance measures.
\majorrevised{In spite of its simplicity, TERAN can outperform current state-of-the-art models on these two retrieval metrics, competing with the currently very effective entangled visual-textual matching models, which on the contrary are not able to produce features for scalable retrieval. Furthermore, we showed that TERAN can successfully output visually-pleasant word-region alignments.}
We also observed that a further reduction of the network complexity can be obtained by sharing the weights of the last TE layers. This has important benefits also on the stability and in the generalization abilities of the whole architecture.

In the end, we think that this work proposes an interesting path towards efficient and effective cross-modal information retrieval.


\begin{acks}
This work was partially supported by “Intelligenza Artificiale per il Monitoraggio Visuale dei Siti Culturali" (AI4CHSites) CNR4C program,
CUP B15J19001040004, by the AI4EU project,
funded by the EC (H2020 - Contract n. 825619), and AI4Media under GA 951911.
%
\end{acks}

\bibliographystyle{ACM-Reference-Format}
\bibliography{biblio}










\end{document}